\documentclass[sigconf]{acmart}
\usepackage{soul}
\usepackage{mathrsfs}
\usepackage{multirow}
\usepackage{hyperref}
\AtBeginDocument{%
  \providecommand\BibTeX{{%
    \normalfont B\kern-0.5em{\scshape i\kern-0.25em b}\kern-0.8em\TeX}}}

\setcopyright{acmcopyright}
\copyrightyear{2022}
\acmYear{2022}
\acmDOI{XXXXXXX.XXXXXXX}

\acmConference[MM '30]{the 30th
	ACM International Conference on Multimedia }{October 10--14,
  2022}{Libon, Portugal}
\acmPrice{15.00}
\acmISBN{978-1-4503-XXXX-X/18/06}

\begin{document}

\title{A Video Anomaly Detection Framework based on Appearance-Motion Semantics Representation Consistency}


\author{Xiangyu Huang}
\affiliation{%
  \institution{Department of Informatics, Xiamen University}
  \city{Xiamen}
  \country{China}}
\email{huangxiangyu@stu.xmu.edu.cn}

\author{Caidan Zhao*}
\affiliation{%
  \institution{Inria Paris-Rocquencourt, Xiamen University}
  \city{Xiamen}
  \country{China}}
\email{zcd@xmu.edu.cn}

\author{Yilin Wang}
\affiliation{%
  \institution{Department of Informatics, Xiamen University}
  \city{Xiamen}
  \country{China}}
\email{wangyilin@stu.xmu.edu.cn}
\author{Zhiqiang Wu}
\affiliation{%
  \institution{College of Engineering, Tibet University\\
  Department of Electrical Engineering, Wright State University}
  \city{}
  \country{}}
\email{zhiqiang.wu@wright.edu}
%
%
%


\begin{abstract}
  Video anomaly detection refers to the identification of events that deviate from the expected behavior. Due to the lack of anomalous samples in training, video anomaly detection becomes a very challenging task. Existing methods almost follow a reconstruction or future frame prediction mode. However, these methods ignore the consistency between appearance and motion information of samples, which limits their anomaly detection performance. Anomalies only occur in the moving foreground of surveillance videos, so the semantics expressed by video frame sequences and optical flow without background information in anomaly detection should be highly consistent and significant for anomaly detection. Based on this idea, we propose Appearance-Motion Semantics Representation Consistency (AMSRC), a framework that uses normal data's appearance and motion semantic representation consistency to handle anomaly detection. Firstly, we design a two-stream encoder to encode the appearance and motion information representations of normal samples and introduce constraints to enhance further the consistency of the feature semantics between appearance and motion information of normal samples so that abnormal samples with low consistency appearance and motion feature representation can be identified. Moreover, the lower consistency of appearance and motion features of anomalous samples can be used to generate predicted frames with larger prediction errors, which makes anomalies easier to spot. Experimental results demonstrate the effectiveness of the proposed method.
\end{abstract}




\maketitle

\section{Introduction}
Video anomaly detection refers to identifying events that do not conform to expected behavior \cite{chandola2009anomaly} in surveillance videos. With the widespread deployment of surveillance cameras in public places recently, video anomaly detection, which is a technology that can interpret the surveillance video content without manual labor, has important application value in public safety scenarios, so it has been appealing to academia. Despite many efforts \cite{luo2017revisit, cong2011sparse, feng2016deep, liu2018future}, video anomaly detection remains an open and very challenging task due to the following two difficulties \cite{chandola2009anomaly}. (1) \emph{Ambiguity}: The forms of abnormal events are unbounded and ambiguous. Since the anomaly has no fixed semantics, the high variability and unpredictability of anomalies make it impossible to model abnormal events directly. (2) \emph{Shortage of anomalies}: The abnormal events usually much less happen than normal ones, so collecting all kinds of anomalous samples is not feasible. It is hard to detect abnormal events based on the above difficulties by training a supervised binary classification model.

Therefore, a typical solution to video anomaly detection is often formulated as an unsupervised learning problem, where the goal is to train a model by using only normal data to mine regular patterns. Then events that do not conform to this model are viewed as anomalies. Based on this scheme, existing methods can be divided into classic hand-crafted feature-based and deep neural network-based methods. Classic video anomaly detection \cite{adam2008robust, benezeth2009abnormal, kim2009observe} needs to manually extract high-level features that can interpret the content of video activities, such as speed and motion trajectory, or low-level features about video frames, such as pixel gradients and textures. Then these extracted features are used to spot anomalies by classic classification methods for anomaly detection, such as a one-class support vector machine. However, feature engineering of such methods is time-consuming and labor-intensive, and the extracted features may be sub-optimal and not robust among other different complex scenarios \cite{xu2017detecting}. With the outstanding achievements of deep neural networks in computer vision tasks, many video anomaly detection methods based on deep neural networks have been proposed and achieved good performance \cite{lu2013abnormal, luo2017revisit, liu2018future, hasan2016learning, lu2019future, luo2017remembering, nguyen2019anomaly, cai2021appearance, liu2021hybrid}.
  
Existing video anomaly detection methods based on deep neural networks almost follow a reconstruction or future frame prediction mode. Reconstruction-based methods \cite{hasan2016learning, luo2017remembering, gong2019memorizing, park2020learning} usually train an autoencoder on normal data and expect abnormal data to incur larger reconstruction errors at test time, making abnormal data detectable from normal ones. Future frame prediction-based methods \cite{liu2018future} use the temporal characteristics of video frames to predict the next frame based on a given sequence of previous frames, then use the prediction errors for anomaly measuring. However, existing studies \cite{gong2019memorizing, zaheer2020old, zong2018deep} have shown that autoencoders trained only on normal data can also reconstruct abnormal ones well, which leads to the poor performance of such methods. Some researches \cite{xu2017detecting, yan2018abnormal, vu2019robust} show that the previous methods neglect to fully utilize motion information of activities. The motion information contains a lot of semantics representing behavioral properties of activities, so modeling motion information is helpful for the detection of abnormal events. However, these methods only combine the information of appearance and motion to detect anomalies in the test phase and do not jointly model the two types of information in the same space during the training phase \cite{cai2021appearance}, which makes it difficult to capture the correlation between the two modalities for anomaly detection. So some novel hybrid methods \cite{cai2021appearance, liu2021hybrid} were proposed to model the consistent correlation between appearance and motion to achieve good performance in video anomaly detection. However, previous methods do not directly model samples' appearance and motion semantic representation consistency to handle anomaly detection. Since anomalies only occur in the foreground of the surveillance video, so the model’s attention should focus on the moving foreground part, rather than the background which is less relevant for behavior. So for the video anomaly detection task, the semantics of appearance and motion features extracted from frame sequences and optical flow without background information should be consistent since the two modalities all represent the foreground behavior properties in the surveillance video. Therefore, modeling the appearance and motion semantic representation consistency of normal samples adequately can make ambiguous anomalies with the lower consistency of two modalities detectable from normal ones.
  
This paper proposes Appearance-Motion Semantics Representation Consistency (AMSRC), a novel framework that adequately models the appearance and motion semantic representation consistency of normal data for video anomaly detection. As illustrated in Figure 1, the two-stream encoder-based future frame prediction model takes both previous video frames and optical flows as input. During the training phase, the two-stream encoder is trained to extract the appearance and motion information representations by only normal samples. And we use consistency constraints to make the feature of two modalities similar, which impel the model to adequately encode the consistent semantics representation between appearance and motion of normal data. We observe that such a well-designed two-stream encoder can encode normal samples to generate the high consistent appearance and motion feature well while producing lower consistent ones for abnormal samples, which can be used to detect anomalies. Moreover, we hope that the difference in appearance-motion semantics representation consistency can lead to the difference in the quality of the predicted frame, thereby further widening the difference between normal and abnormal samples. So we propose a gated fusion module for fusing the features generated by the two-stream encoder. Inconsistent representations between appearance and motion are activated to produce a feature quite different from the ones before fusion. The above design facilitates to utilize the feature semantics consistency gap between normal and abnormal data to augment the quality of the prediction gap. Normal data will generate consistent appearance-motion features, which are trained to predict the future frame with higher quality. In contrast, lower consistent appearance-motion features generated by abnormal data will produce an irregular feature through the gated fusion module, and the irregular feature will produce future frames with larger prediction errors. So during the anomaly detection phase, we use the semantics consistency of appearance-motion features and the frame prediction errors as final video anomaly detection cues.
  
We summarize our contributions as follows: 
\begin{itemize}
  \item We propose AMSRC (Appearance-Motion Semantics Representation Consistency), a framework that uses the appearance and motion semantic representation consistency gap between normal and abnormal data to spot anomalies. 
  \item We introduce a gated fusion module so that the appearance-motion feature semantics inconsistency will lead to low quality of the predicted frame, to a certain extent, ensuring that abnormal samples can generate larger prediction errors on autoencoders trained only with normal data.
  \item Extensive experiments on three standard public video anomaly detection datasets demonstrate our methods’ effectiveness, and all code will be released for further research convenience to the community.
\end{itemize}
\begin{figure*}[ht]
	\centering
	\includegraphics[width=14cm]{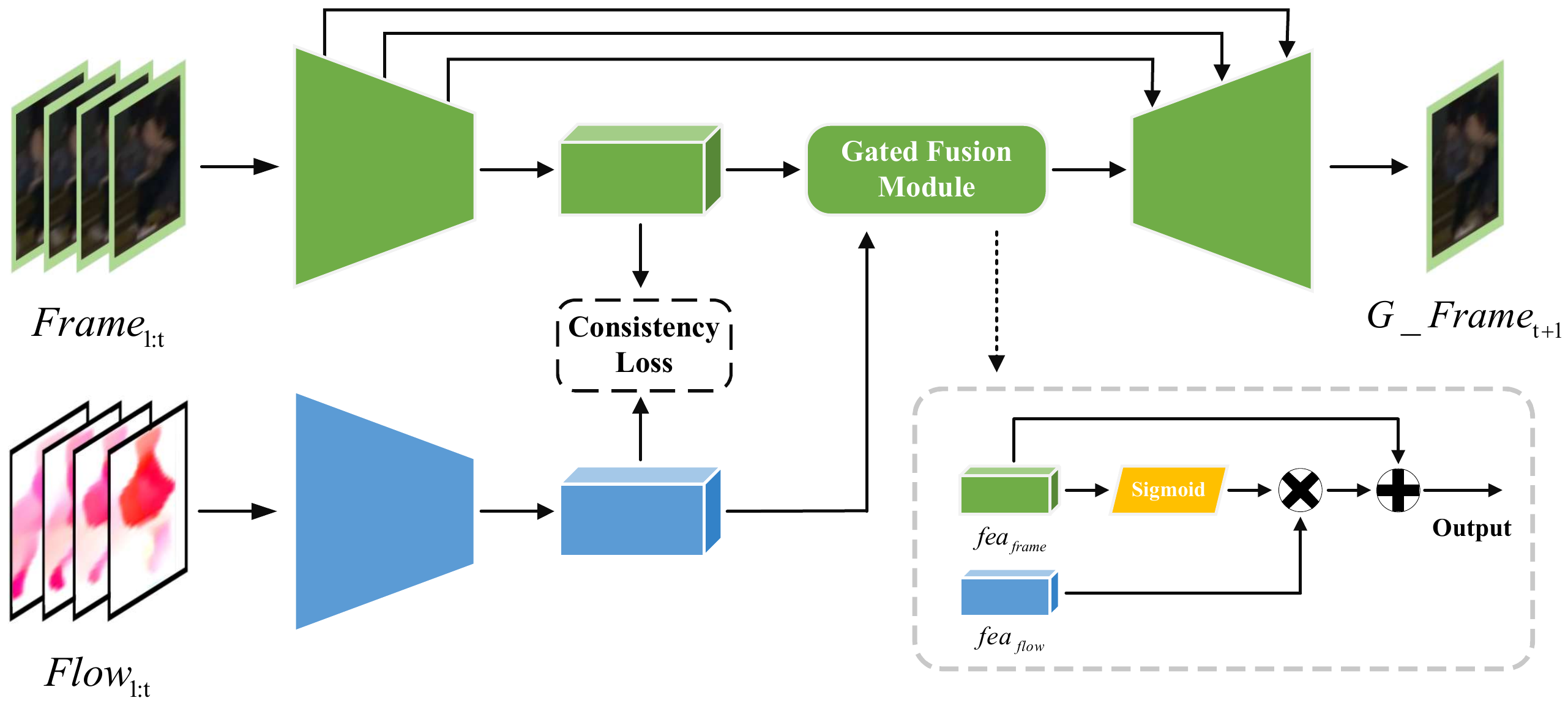}
	\caption{Overview of the proposed Appearance-Motion Semantics Representation Consistency Network (AMSRC-Net). Our model takes a sequence of previous frame images and the corresponding optical flows as the inputs. During the training phase, our model is trained to encode frames and optical flows information, and we use the consistency constraints to increase the similarity of frames and optical flows features at the bottleneck layer of the two-stream encoder. Then, the features of frames and flows are fused by a gated fusion module to predict the next future frame. If an abnormal event occurs during the testing phase, lower consistent appearance-motion features are generated. And these lower consistent appearance-motion will be fused by the gated fusion module to guide future frame prediction. The prediction errors will be enlarged further.}
	\label{p1}
\end{figure*}
\section{Related Work}
Recently, many researches have been done on video anomaly detection, and a large number of methods have been proposed to solve this difficulty. Existing methods can be divided into classic hand-crafted feature-based and deep neural network-based methods. Classic hand-crafted feature-based methods are mainly comprised of two stages: Feature extraction by hand-crafted descriptors for video content and anomaly detection by classic one-class machine learning methods. Early work typically uses low-level trajectory features, such as image coordinates, to represent regular patterns \cite{tung2011goal, wu2010chaotic}. Since the trajectory features are based on object tracking, these methods are not suitable for complex or crowded scenes. So more low-level features are proposed for anomaly detection, such as histogram of oriented flows \cite{cong2011sparse}, spatio-temporal gradients \cite{lu2013abnormal, kratz2009anomaly}, and dynamic texture \cite{mahadevan2010anomaly}. Moreover, various machine learning methods for video anomaly detection, such as probabilistic models \cite{kim2009observe, mahadevan2010anomaly}, sparse coding \cite{cong2011sparse, lu2013abnormal}, and one-class classifier \cite{yin2008sensor}. have been widely studied. However, feature engineering of such methods is time-consuming and labor-intensive. Due to the limited representation capability of the designed descriptors, it is hard to ensure the robustness of the methods across different complex scenarios. Benefiting from the powerful representation capabilities of Convolutional Neural Networks (CNNs), a large number of deep learning-based anomaly detection methods have been proposed. And the anomaly detection mode based on frame reconstruction or future frame prediction is the current mainstream method, which shows strong detection performance. In the frame reconstruction-based paradigm, autoencoders and their variants are widely proposed to reconstruct the training data, such as ConvAE \cite{hasan2016learning} and ConvLSTM-AE \cite{luo2017remembering}. These methods assume that an autoencoder trained only on normal data cannot reconstruct abnormal ones well. However, this assumption does not always hold, and the autoencoder sometimes can also reconstruct anomalous data well \cite{gong2019memorizing, zaheer2020old}. To avoid this problem, Liu et al. \cite{liu2018future} first proposed a paradigm based on future frame prediction, which uses the future frames’ prediction errors as an anomaly indicator. While this paradigm has strong performance, its validity is still based on the assumption that anomalies are usually unpredictable. Furthermore, some works hope to take full advantage of both paradigms and combine the two paradigms to develop hybrid approaches. In \cite{nguyen2019anomaly}, Nguyen et al. proposed an autoencoder consisting of a shared encoder and two separate decoders for frame reconstruction and optical flow prediction. Ye et al. \cite{ye2019anopcn} decomposes the reconstruction paradigm into prediction and refinement, then proposed a predictive coding network. Liu et al. \cite{liu2021hybrid} seamlessly combine optical flow reconstruction and frame prediction so that the error of flow reconstruction can affect the results of frame prediction. Anomalies that generate flow reconstruction error will deteriorate the quality of anomalous predicted frames so that anomalies can be easily detected. However, the previous method ignored the consistent correlation between appearance and motion information representation in video anomaly detection. Cai et al. \cite{cai2021appearance} proposed an appearance-motion memory consistency network (AMMC-Net) to model the appearance-motion correspondence in high-level feature space. AMMC-Net encodes the video frames and the corresponding optical flows and uses a memory network for storage. Then the encoding got by the memory network, and the actual frame features are combined to generate the final representations. However, such a network’s performance is highly dependent on the memory size, and a small-sized memory network may seriously limit normal data reconstruction capability in complex scenes. Compared to the above method, our work directly encodes the corresponding semantic representation of appearance-motion for activities on the moving foreground and uses a simple gated fusion module to make the inconsistency representations between appearance and motion of anomalies affect the quality of the prediction frame, so that abnormal samples can produce larger prediction errors to make anomalies easier to spot.

\section{Proposed Method}
\begin{figure*}[ht]
	\centering
	\includegraphics[width=18cm]{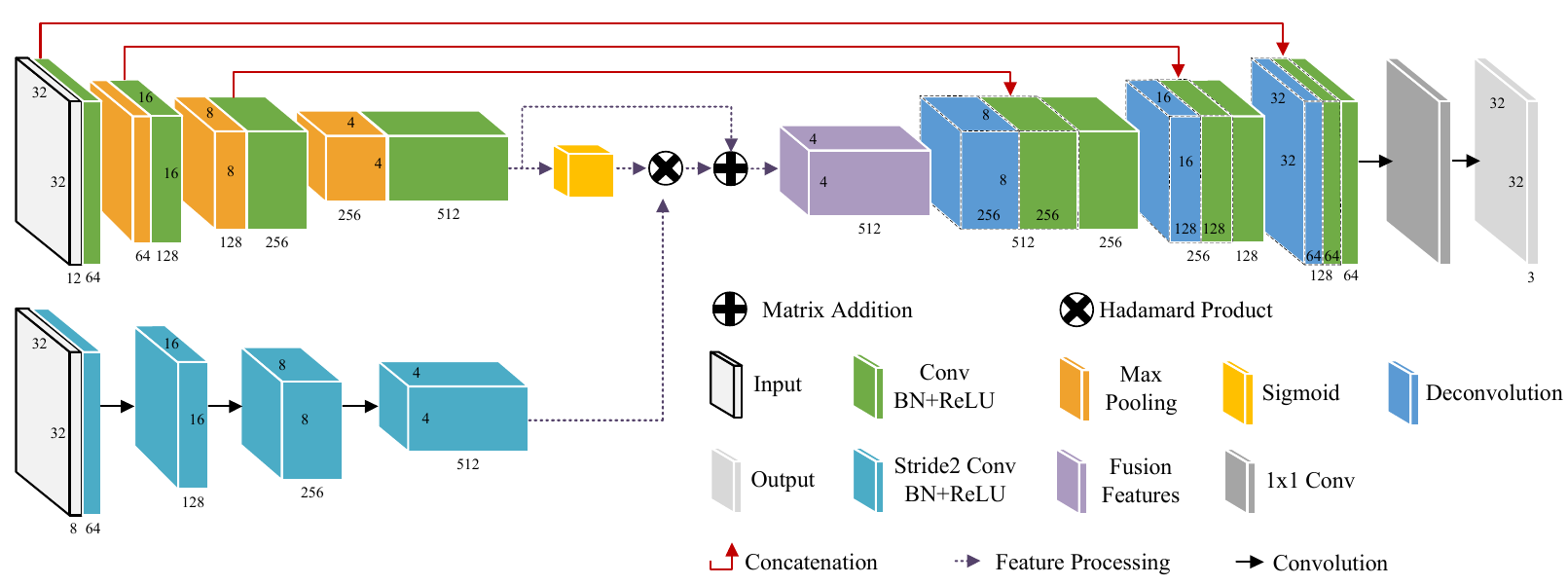}
	\caption{Detailed network architecture of AMSRC-Net in our experiments.}
	\label{p2}
\end{figure*}

As shown in Figure \ref{p1}, our proposed AMSRC-Net consists of three parts:  A two-stream encoder, a decoder, and a gated fusion module. We first input a previous video frame image and its optical flow clip into the two-stream encoder to get the appearance and motion’s feature representations. Then we add constraints to further enhance the consistency of the feature semantics between appearance and motion information of normal samples. Next, two consistent modalities features are input into the gated fusion module. Finally, feeding the fused feature into the decoder to predict the future frame image. The detailed network architecture of AMSRC is shown in Figure 2.
\subsection{Two-stream Encoder and Decoder}
The two-stream encoder extracts feature representations from input video frame images and the corresponding optical flows. Due to the consistency constraints, the extracted features’ semantics are highly similar, representing the foreground behavior properties in the surveillance video. Then the decoder is trained to generate the next frame by taking the aggregated feature formed by fusing the extracted features from the previous step. While the aggregated feature maybe lacks low-level information, such as backgrounds, textures, and so on. To solve this problem, we add a UNet-like skip connection structure \cite{ronneberger2015u} between the frame stream encoder and decoder to preserve these low-level features irrelevant to behavior for predicting the high-quality future frame.
\begin{figure}[hb]
	\centering
	\includegraphics[width=8cm]{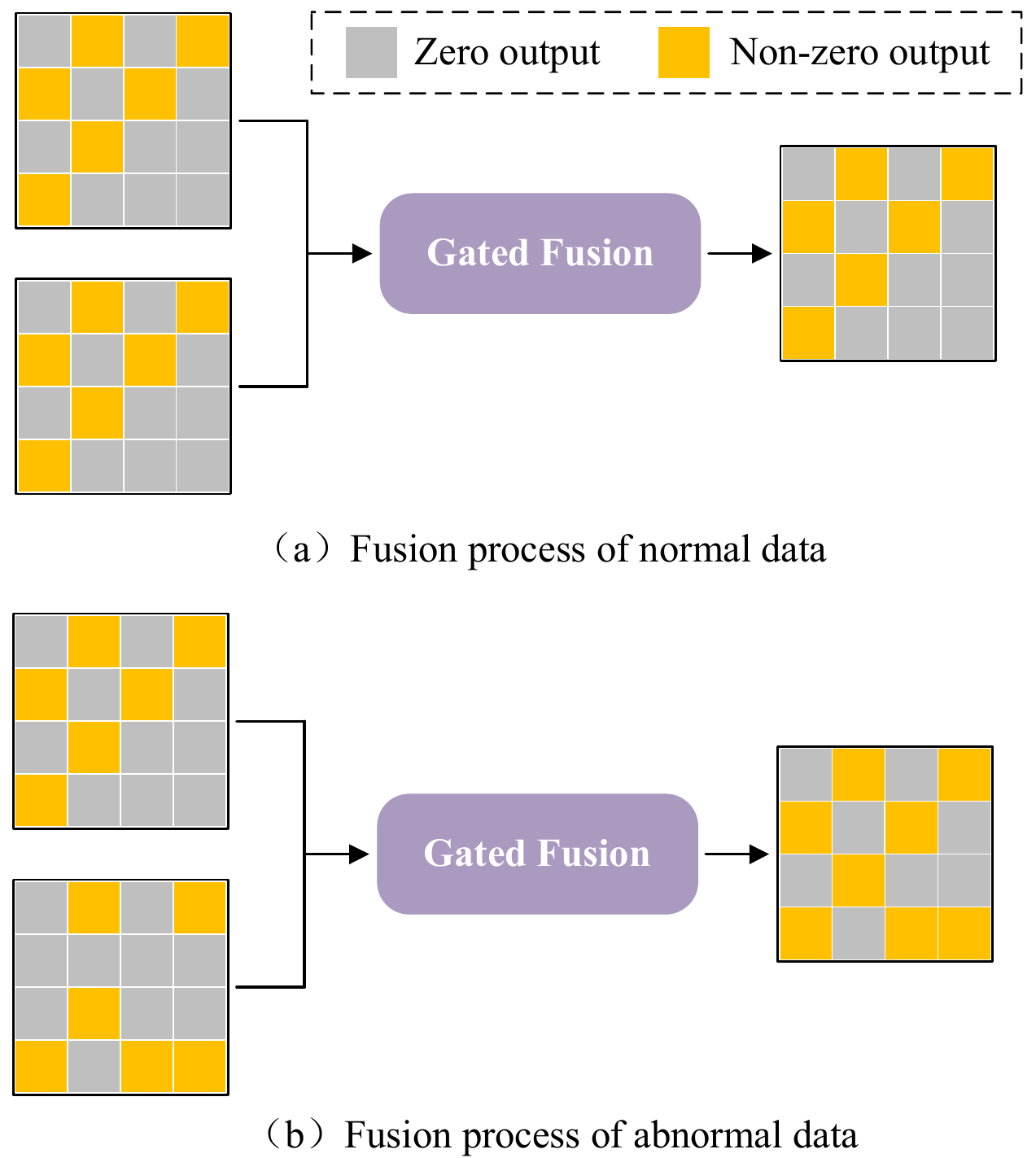}
	\caption{The visualization of appearance and motion features of samples passing through the gated fusion module: (a) fusion process of appearance an motion features of normal data, (b) fusion process of appearance an motion features of abnormal data.}
	\label{p3}
\end{figure}
\subsection{Gated Fusion Module}
Since Relu activation is adopted at the end of the two-stream encoder, there are many feature representations with a value of zero in the output features. Based on the previous consistent appearance-motion representation constraints, we observe that the appearance and motion feature representations with a zero value are highly similar in distribution. In contrast, the lower consistency of appearance-motion features generated by abnormal data reflects a larger difference in the distribution of the appearance and motion feature representations with a value of zero. In order to utilize this feature representation gap to improve the anomaly detection performance further, we aim to design a gated fusion mechanism to generate a different representation between normal and abnormal samples.

The structure of our proposed gated fusion module is shown in Figure \ref{p1}. The gated fusion module uses Sigmoid activation to deal with the feature of frame images, the appearance feature representations with a value of zero will be reactivated as output between 0 and 1. Then we multiply the activated output by the feature of the corresponding optical flows, preserving the conflicting information between appearance and motion features. Finally, we add the feature of frame images to the previous result as an input for the decoder to generate the predicted frame. So due to inconsistent appearance and motion feature of anomalies, the gated fusion module will generate a feature that is different from the pre-fusion representation. Thus, the model will produce larger prediction errors for abnormal samples to improve anomaly detection performance. The visualization of the appearance and motion features of samples passing through the gated fusion module is shown in Figure \ref{p3}.

\subsection{Loss Function}
We follow the previous anomaly detection work based on future frame prediction \cite{liu2018future}, using intensity and gradient difference to make the prediction close to its ground truth. The intensity loss guarantees the similarity of pixels between the prediction and its ground truth, and the gradient loss can sharpen the predicted images. Specifically, we minimize the $\ell_{2}$ distance between the predicted frame $\hat{x}$ and its ground truth $x$ as follows:
\begin{equation}
	L_{i n t}=\left\|\hat{x}-x\right\|_{2}^{2}
\end{equation}

The gradient loss is defined as follows:
\begin{equation}
	\begin{aligned}
		L_{g d}=\sum_{i, j}&\left\|\left|x_{i, j}-x_{i-1, j}\right|-\left|x_{i, j}-x_{i-1, j}\right|\right\|_{1}\\
		+&\left\|\left|x_{i, j}-x_{i, j-1}\right|-\left|x_{i, j}-x_{i, j-1}\right|\right\|_{1}
	\end{aligned}
\end{equation}
where $i$, $j$ denote the spatial index of a video frame.

In order to model the appearance and motion semantic representation consistency of normal samples, we minimize the cosine distance between appearance and motion features of normal samples encoded by the two-steam encoder. So the consistency loss is defined as follows:
\begin{equation}
	L_{sim}=1-\frac{\langle{fea_{frame}}, fea_{flow}\rangle}{\|fea_{frame}\left\|_{2}\right\|fea_{flow}\|_{2}}
\end{equation}
where $fea_{frame}$, $fea_{flow}$ denote the appearance and motion feature encoded by the two-steam encoder, respectively.

Then, the overall loss $L$ for training takes the form as follows:
\begin{equation}
	L=\lambda_{int} L_{i n t}+\lambda_{g d} L_{g d}+ \lambda_{sim} L_{sim}+\lambda_{model} \left\|W\right\|_{2}^{2}
\end{equation}
where $\lambda_{int}$, $\lambda_{gd}$, and $\lambda_{sim}$ are balancing hyper-parameters, $W$ is the parameter of the model, and $\lambda_{model}$ is a regularization hyper-parameter that controls the model complexity. 
\subsection{Anomaly Detection}
Our anomaly score is composed of two parts during the testing phase: the inconsistency of appearance and motion feature $S_{f}=1-\frac{\langle{fea_{frame}}, fea_{flow}\rangle}{\|fea_{frame}\left\|_{2}\right\|fea_{flow}\|_{2}}$ and the future frame prediction error $S_{p}=\left\|\hat{x}-x\right\|_{2}^{2}$. Then, we get the final anomaly score by fusing the two parts using a weighted sum strategy as follows:
\begin{equation} \label{e5}
	\mathrm{S}=w_{f} \frac{S_{f}-u_{f}}{\delta_{f}}+w_{p} \frac{S_{p}-u_{p}}{\delta_{p}}
\end{equation}
where $u_{f}$, $\delta_{f}$, $u_{p}$, and $\delta_{p}$ denote means and standard deviations of the inconsistency between appearance and motion feature and prediction error of all the normal training samples. $w_{f}$ and $w_{p}$ represent the weights of the two scores.
\begin{table*}
	\caption{AUROC (\%) comparison between the proposed AMSRC and state-of- the-art video anomaly detection methods on UCSD ped2, CUHK Avenue and ShanghaiTech datasets.}
	\setlength{\tabcolsep}{7mm}{
		\begin{tabular}{ccccc}
			\toprule[1.5pt]
			\multicolumn{2}{c}{Method}                                                                                                & USCD Ped2     & CUHK Avenune  & ShanghaiTech  \\ \midrule[1.5pt]
			\multirow{3}{*}{\begin{tabular}[c]{@{}c@{}}Classic Video Anomaly \\ Detection Methods\end{tabular}} & MPPCA\cite{kim2009observe}       & 69.3          & N/A           & N/A           \\
			& MPPC+SFA\cite{mahadevan2010anomaly}     & 61.3          & N/A           & N/A           \\
			& MDT\cite{mahadevan2010anomaly}          & 82.9          & N/A           & N/A           \\ \midrule[1pt]
			\multirow{4}{*}{\begin{tabular}[c]{@{}c@{}}Reconstruction-Based \\ Methods\end{tabular}}            & ConvAE\cite{hasan2016learning}      & 90            & 70.2          & N/A           \\
			& ConvLSTM-AE\cite{luo2017remembering} & 88.1          & 77            & N/A           \\
			& MemAE\cite{gong2019memorizing}       & 94.1          & 83.3          & 71.2          \\
			& MNAD-R\cite{park2020learning}      & 90.2          & 82.8          & 69.8          \\ \midrule[1pt]
			\multirow{3}{*}{\begin{tabular}[c]{@{}c@{}}Prediction-Based \\ Methods\end{tabular}}                & Frame-Pred.\cite{liu2018future}  & 95.4          & 85.1          & 72.8          \\
			& MNAD-R\cite{park2020learning}      & 97            & 88.5          & 70.5          \\
			& VEC\cite{yu2020cloze}         & 97.3          & 90.2          & 74.8          \\ \midrule[1pt]
			\multirow{5}{*}{\begin{tabular}[c]{@{}c@{}}Hybrid and Other \\ Methods\end{tabular}}                & Stacked RNN\cite{luo2017revisit}  & 92.2          & 81.7          & 68            \\
			& AMC\cite{nguyen2019anomaly}         & 96.2          & 86.9          & N/A           \\
			& AnoPCN\cite{ye2019anopcn}      & 96.8          & 86.2          & 73.6          \\
			& AMMC-Net\cite{cai2021appearance}    & 96.6          & 86.6          & 73.7          \\
			& $\text{HF}^{2}$-VAD\cite{liu2021hybrid}     & 99.3          & 91.1          & 76.2          \\ \midrule[1pt]
			\textbf{Proposed}                                                                                            & AMSRC         & \textbf{99.3} & \textbf{93.8} & \textbf{76.3} \\ \bottomrule[1.5pt]
	\end{tabular}}
	\label{t1}
\end{table*}
\section{Experiments}

\subsection{Datasets}
\begin{figure}[ht]
	\centering
	\includegraphics[width=8cm]{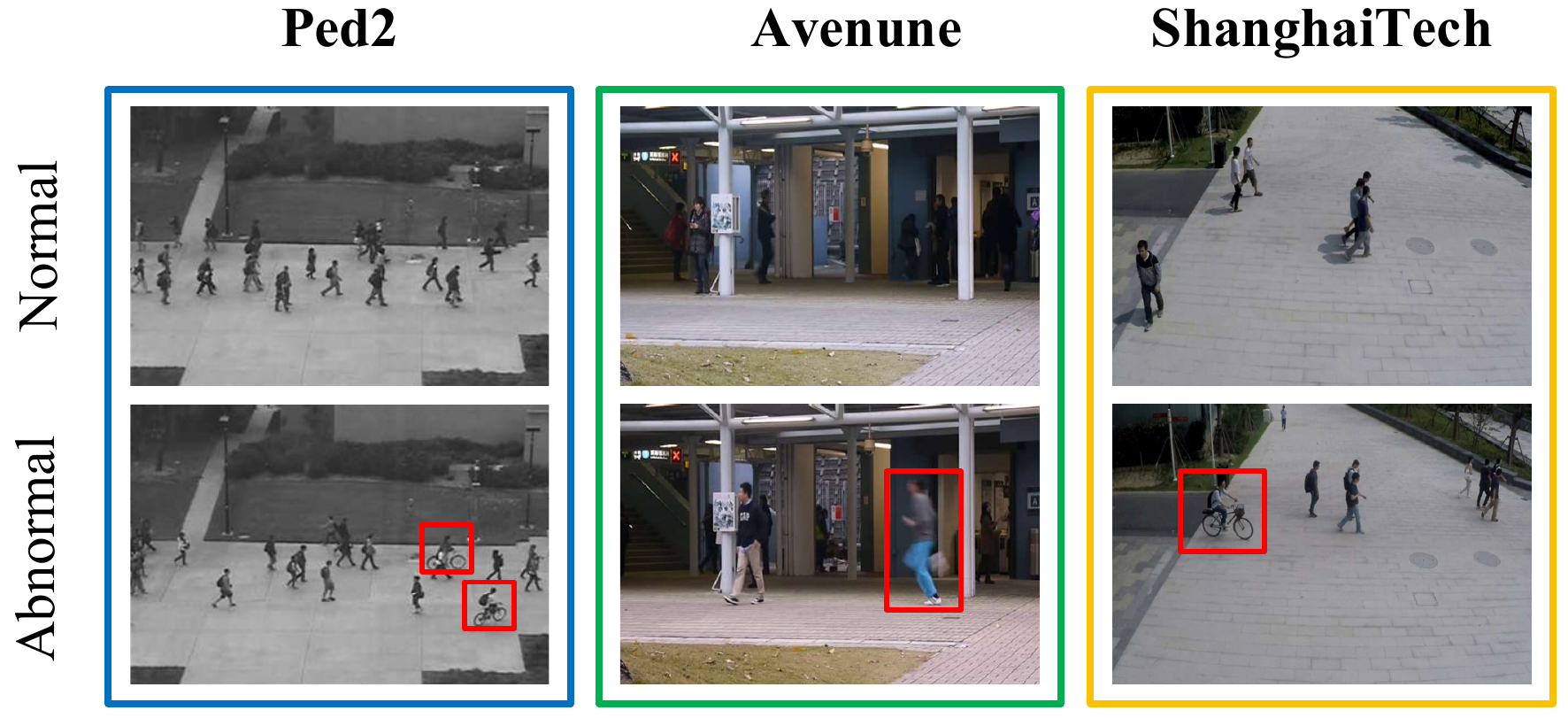}
	\caption{Some examples including normal and abnormal frames in the UCSD ped2, CUHK Avenue and ShanghaiTech datasets are shown. Red boxes denote anomalies happened in abnormal frames.}
	\label{p4}
\end{figure}

We evaluate our approach on three standard popular video anomaly detection datasets, including UCSD ped2 \cite{mahadevan2010anomaly}, CUHK Avenue \cite{lu2013abnormal}, and ShanghaiTech \cite{luo2017revisit}. Some samples are shown in Figure \ref{p4}.
\begin{itemize}
	\item UCSD ped2 dataset contains 16 training videos and 12 testing videos with 12 abnormal events, acquired with a stationary camera. The normal training data consists of only pedestrians walking, while the anomalies in testing videos are the appearance of the non-pedestrian objects (\emph{e.g.} vehicles and cars) and strange pedestrian motion.
	\item CUHK Avenue dataset contains 16 training videos and 21 testing videos with 47 abnormal events collected from a fixed scene. The anomalies in testing videos consist of running, throwing bags, and loitering. Especially, the size of people may change due to the camera position and angle.
	\item ShanghaiTech dataset is a very challenging dataset for video anomaly detection. It consists of 330 training videos and 107 testing ones with 130 abnormal events collected from 13 scenes with different camera positions and angles. The anomalies contain the non-pedestrian objects (\emph{e.g.} specialsymbols vehicles) and strange motion (\emph{e.g.} chasing and fighting).
\end{itemize}
\subsection{Evaluation Criterion}
We follow the widely popular evaluation metric in video anomaly detection \cite{liu2018future, cai2021appearance, liu2021hybrid} and evaluate our method using the frame-level area under the ROC curve (AUC) metric. The ROC curve is measured by varying the threshold over the anomaly score. Higher AUC values represent better performance for anomaly detection.

\subsection{Parameters and Implementation Details}
Following \cite{liu2021hybrid, yu2020cloze}, we train our model on the patches with foreground objects instead of the whole video frames. In advance, all foreground objects are extracted from original videos for the training and testing samples. RoI bounding boxes identify foreground objects. For each RoI, a spatial-temporal cube (STC) \cite{yu2020cloze} composed of the object in the current frame and the content in the same region of previous t frames will be built, where the hyper-parameter t is set to 4. And the width and height of STCs are resized to 32 pixels. The corresponding optical flows are generated by FlowNet2 \cite{ilg2017flownet}, and the STCs for optical flows are built in a similar way. Due to existing many objects in a frame, we select the maximum anomaly score of all objects as the anomaly score of a frame.

The implementation of our AMSRC is done in PyTorch \cite{paszke2019pytorch}, and we adopt Adam optimizer \cite{kingma2014adam} to optimize it. The initial learning rate is set to $2e^{-4}$, decayed by 0.8 after every ten epochs. The batch size and epoch number of Ped2, Avenue, and ShanghaiTech are set to $(128, 60)$, $(128, 40)$, $(256, 40)$. $\lambda_{int}$, $\lambda_{gd}$, $\lambda_{sim}$, and $\lambda_{model}$ for Ped2, Avenue, and ShanghaiTech are set to $(1, 1, 1, 1)$, $(1, 1, 1, 1)$, $(1, 1, 10, 1)$. Then the error fusing weights $w_{f}$, $w_{p}$ for Ped2, Avenue, and ShanghaiTech are set to $(1, 0.01)$, $(0.2, 0.8), (0.4, 0.6)$. All experiments are done on an NVIDIA RTX 3090 GPU and an intel XEON GOLD 6130 CPU $@$ 2.1GHz.

\subsection{Anomaly Detection Results}
\begin{figure}[ht]
	\centering
	\includegraphics[width=8.5cm]{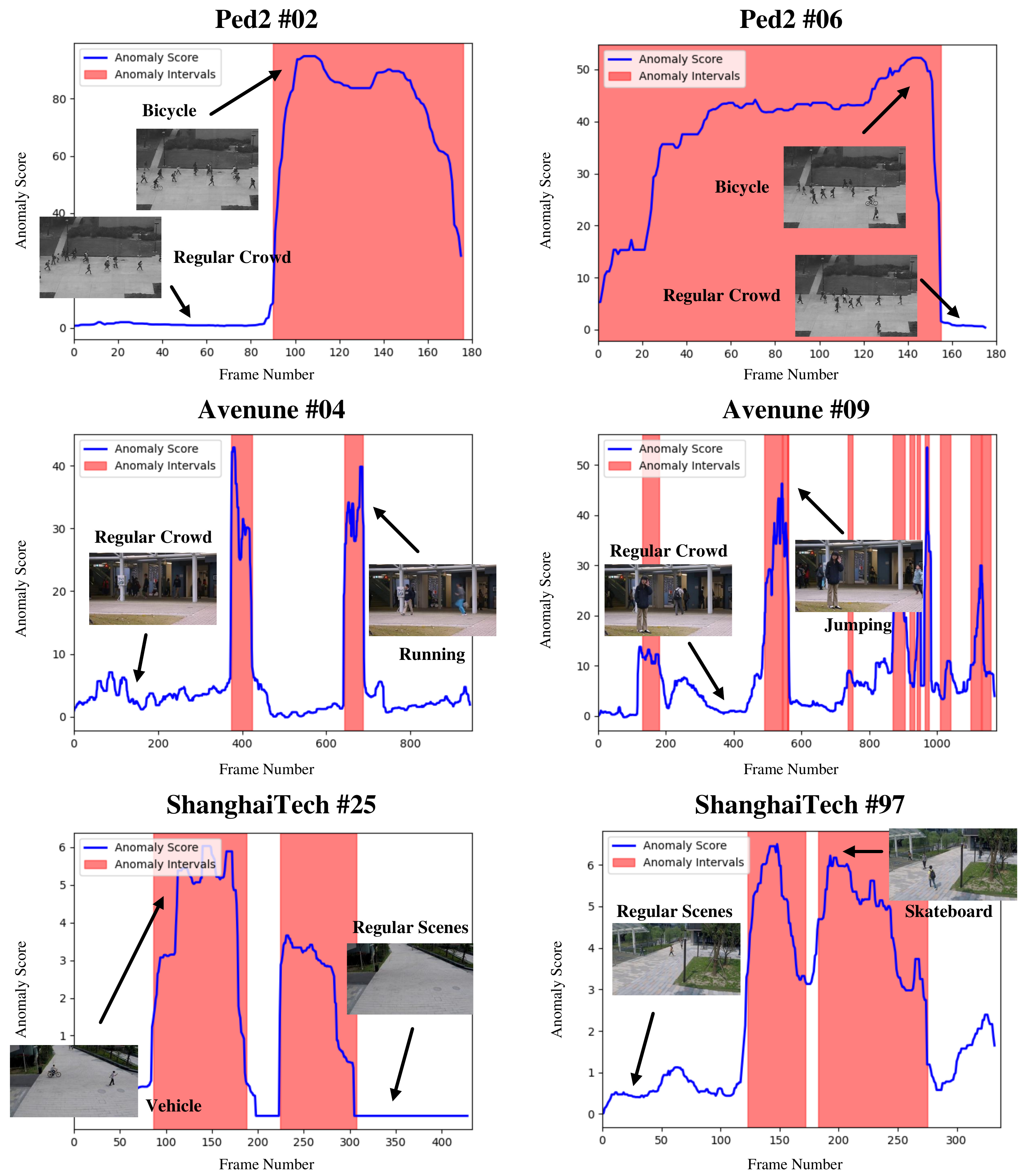}
	\caption{Anomaly score curves of six testing video clips from Ped2, Avenue, ShanghaiTech datasets. Red regions represent ground truth anomalous frames. It shows that the anomaly scores rise when abnormal events occur. Some examples are sampled to display the normal/abnormal events in each scenario. Best viewed in color.}
	\label{p5}
\end{figure}

To evaluate the performance of our AMSRC, anomaly detection is performed on three popular standard datasets. Examples in Figure \ref{p5} show anomaly score curves of six testing video clips from Ped2, Avenue, and ShanghaiTech datasets. The anomaly score is calculated by Equation \ref{e5} and can be utilized to detect anomalies. The red regions denote the ground truth anomalous frames. As can be seen, the anomaly score of a video clip rises when anomalies occur, and descents when anomalies disappear, which shows our method can spot the anomalies accurately.

Within our best knowledge, we compare our AMSRC with state-of-the-art methods, including: (1) classic video anomaly detection methods: MPPCA \cite{kim2009observe}, MPPC+SFA \cite{mahadevan2010anomaly}, and MDT \cite{mahadevan2010anomaly}; (2) reconstruction-based methods: ConvAE \cite{hasan2016learning}, ConvLSTM-AE \cite{luo2017remembering}, MemAE \cite{gong2019memorizing}, and MNAD-R \cite{park2020learning}; (3) prediction-based methods: Frame-Pred \cite{liu2018future}, MNAD-P \cite{park2020learning}, and VEC \cite{yu2020cloze}; (4) hybrid and other methods: Stacked RNN \cite{luo2017revisit}, AMC \cite{nguyen2019anomaly}, AnoPCN \cite{ye2019anopcn}, AMMC-Net \cite{cai2021appearance}, and HF2-VAD \cite{liu2021hybrid}. The results are summarized in Table \ref{t1}, and the performances of compared methods are obtained from their original papers.

As observed, our proposed AMSRC outperforms compared state-of-the-art video anomaly detection methods on three popular standard datasets, demonstrating our method's effectiveness. Especially, AMSRC outperforms AMMC-Net, which also models the appearance-motion correspondence. And we observe that the methods which use memory networks to keep the representations, such as MemAE \cite{gong2019memorizing}, MNAD-R \cite{park2020learning}, MNAD-P \cite{park2020learning}, and AMMC-Net \cite{cai2021appearance}, have a limited performance for anomaly detection on Avenue and ShanghaiTech. The difficulty of modeling a suitable-sized memory network will limit their performance on Avenue and ShanghaiTech, which contain complex scenes and abnormal events. While we directly model the corresponding semantic representation of appearance-motion to get better performance. In particular, we note that our method achieves 93.8\% frame-level AUROC on CUHK Avenue, which is the best performance achieved on Avenue currently.
\begin{table}[htbp]
	\centering
	\caption{Ablation study result on UCSD ped2, CUHK Avenue and ShanghaiTech datasets. The anomaly detection performance is shown in terms of AUROC (\%).}
	\begin{tabular}{cccc}
		\toprule[1.5pt]
		\multicolumn{1}{c}{\multirow{2}{*}{}} 	    & USCD & CUHK &
		\multicolumn{1}{c}{\multirow{2}{*}{ShanghaiTech}} \\
		                                  & Ped2  & Avenune &  \\
		\midrule[1.5pt]
		Baseline                          & 92.9  & 90.6  & 74.7 \\
		AMSRC w/o & \multirow{2}{*}{97.6} & \multirow{2}{*}{92.5} & \multirow{2}{*}{75.2} \\
		gated fusion module 			  &       &       &  \\
		AMSRC 				              & 99.3  & 93.8  & 76.3 \\
		\bottomrule[1.5pt]
	\end{tabular}
	\label{t2}
\end{table}

\subsection{Detailed Analysis}
\subsubsection{Ablation Studies}
To analyze the role of different components of AMSRC, we perform corresponding ablation studies and display the results in Table \ref{t2}. To evaluate the effectiveness of modeling appearance-motion semantics representation consistency, we only utilize the frame stream encoder and decoder to predict future frames for anomaly detection as a baseline, and the baseline can get 92.9\%, 90.6\%, 74.7\% AUROC scores on UCSD ped2, CUHK Avenue and ShanghaiTech, respectively. Then, we conduct experiments to evaluate the performance of AMSRC without a gated fusion module, modeling the corresponding appearance-motion representations brings evident improvement by 4.7\%, 1.9\%, and 0.5\% AUROC gain on UCSD ped2, CUHK Avenue, and ShanghaiTech respectively. Finally, AMSRC with a gated fusion module obtains 99.3\%, 93.8\%, and 76.3\%. Compared to baseline, the AUROC scores have been improved by 6.4\%, 3.2\%, and 1.6\% on UCSD ped2, CUHK Avenue, and ShanghaiTech, respectively.
\subsubsection{Visulization}
\begin{figure}[ht]
	\centering
	\includegraphics[width=8.5cm]{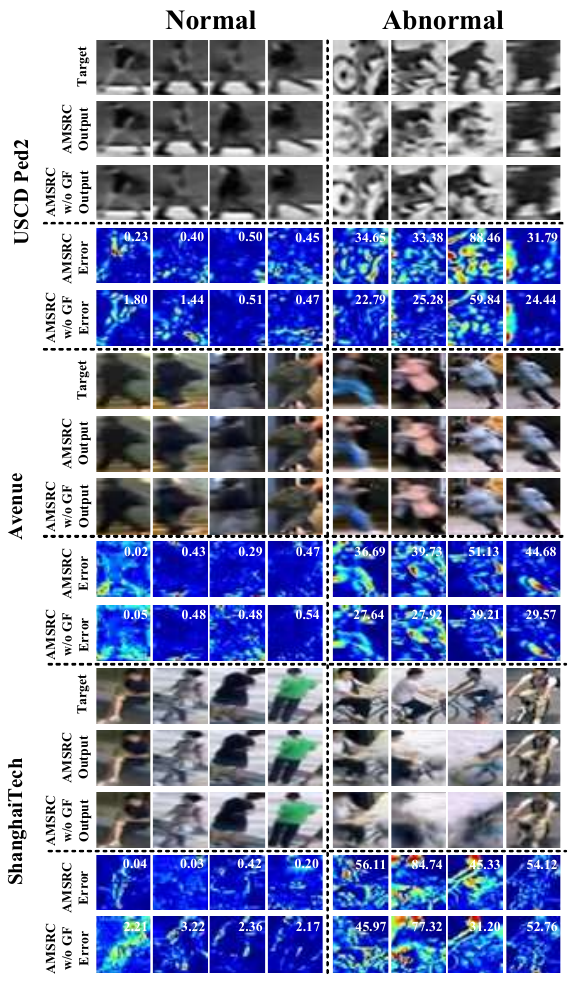}
	\caption{Visualization examples of the ground truth frames (Target), completed frames by AMSRC (AMSRC Output), completed frames by AMSRC without gated fusion module (AMSRC w/o GF Output), completion errors by AMSRC (AMSRC Error), and completion errors by AMSRC without gated fusion module (AMSRC w/o GF Error). Brighter colors in the error map indicate a larger prediction error. Best viewed in color.}
	\label{p6}
\end{figure}

To show that our proposed gated fusion module can help produce larger prediction error for anomalies, we demonstrate the visualized results of representative normal/abnormal events sampled from three popular standard datasets in Figure \ref{p6}. As we can see, AMSRC produces minor differences from normal images. While abnormal events produce large differences, these differences are observed in regions with the motion behavior semantics. Such observations imply that AMSRC pays more attention to high-level behavior semantics for anomalies. Moreover, compared with AMSRC without a gated fusion module, AMSRC produces larger prediction errors for anomalies, which demonstrates the effectiveness of our proposed gated fusion module for anomaly detection.
\section{Conclusion}
In this paper, based on the idea that the semantics of appearance and motion information representations should be consistent, we model the appearance and motion semantic representation consistency of normal data to handle anomaly detection. We design a two-stream encoder to encode the appearance and motion information representations of normal samples and add constraints to strengthen the consistent semantics between appearance and motion information of normal samples so that abnormal ones with lower consistent appearance and motion features can be identified. And the lower consistency of appearance and motion features of anomalies can be fused by our designed gated fusion module to affect the quality of predicted frames, making anomalies produce larger prediction errors. Experimental results on three popular standard datasets show that our method performs better than state-of-the-art approaches.

\bibliographystyle{ACM-Reference-Format}
\bibliography{sample-base}

%
%
%
%
%
%
%
%

\end{document}